%% file: main.tex
\documentclass{article}
\pdfoutput=1

\usepackage{xr}
\usepackage{printlen}

\usepackage{microtype}
\usepackage{graphicx}
\usepackage{amsmath}
\usepackage{mathtools}
\usepackage{nccmath}
\usepackage{xfrac}
\usepackage{booktabs}
\usepackage{siunitx}
\usepackage{natbib}
\renewcommand{\cite}[1]{\citep{#1}}
\usepackage[dvipsnames]{xcolor}

\usepackage{pgfplots}
\usepgfplotslibrary{groupplots}
\pgfplotsset{compat=newest}
\usepackage{tikzexternal}  
\newlength\figurewidth
\newlength\figureheight

\input{setup.tex}
\input{abbrev.tex}

\newcommand*{\myparagraph}[1]{\paragraph{#1}}

\colorlet{dataterm}{green!60!black}
\colorlet{areaterm}{blue}
\colorlet{likelihood}{red!50!blue}

\crefname{figure}{Figure}{Figures}

\allowdisplaybreaks

\begin{document}

\title{Large-Scale Cox Process Inference using Variational Fourier Features}
\author{ST John and James Hensman}
\maketitle

\begin{abstract}
    Gaussian process modulated Poisson processes provide a flexible framework
    for modelling spatiotemporal point patterns. So far this had been
    restricted to one dimension, binning to a pre-determined grid, or small
    data sets of up to a few thousand data points.  Here we introduce Cox
    process inference based on Fourier features.
    This sparse representation induces global rather than local constraints on the function space
    and is computationally efficient.  This allows us to
    formulate a grid-free approximation that scales well with the number of data points
    and the size of the domain.  We demonstrate that this allows MCMC
    approximations to the non-Gaussian posterior.  We also find that, in
    practice, Fourier features have more consistent optimization behavior than
    previous approaches.  Our approximate Bayesian method 
    can fit over \num{100 000} events with complex
    spatiotemporal patterns in three dimensions on a
    single GPU.
\end{abstract}


\section{Introduction}

Modelling spatiotemporal point patterns is a common task in geostatistics, for example in ecology and epidemiology \citep{DiggleTaylorGeostatistics,vanhatalo2007sparse}.
We are further motivated by modelling events occurring in a city, \eg taxi pickups or crime incidents \citep{flaxman2018crime}.

Gaussian process (GP) modulated Poisson processes provide a flexible Bayesian model for such data. 
The model includes a GP prior over a latent function $f(\cdot)$, which is related to the rate $\lambda(\cdot)$ of an inhomogenous Poisson process through a link function. This is usually taken to be $f(\cdot) = \log( \lambda(\cdot))$, resulting in the Log Gaussian Cox process \cite{LGCP}.
These models are computationally challenging because they are doubly intractable \cite{MurrayThesis}. The likelihood involves an integral of the process over the spatiotemporal domain, which cannot be computed in general. There are three potential remedies to this issue.
First, the classic approach is to grid the input domain \citep[see, e.g.,][]{TaylorDiggleGrids}, assuming that the rate is constant over each grid cell. This can be solved with, \eg MCMC.
However, this imposes a discretised structure and scales poorly with the number of grid cells.
A second approach is to use a thinning strategy to construct an exact MCMC sampler \cite{AdamsMurrayMacKay,Gunter2014}, which demands that the inverse link function be bounded, \eg sigmoidal. However, this is prohibitively expensive for more than a few thousand points.
Finally, we can use a square-root link function $\lambda(\cdot) = f(\cdot)^2$, as proposed by \citet{Lloyd2015}. Combined with variational inference for the latent function, this can make the integral tractable. Though their approach scales linearly with the number of events, it scales poorly with the size of the domain, especially if the rate has high variance.
\citet{Flaxman} make use of the same link function with a frequentist approach.

In this work we build on Lloyd et al's proposal. We similarly make use of the square-root link function for tractability, but introduce several innovations which greatly improve the applicability of the method.

First, we extend the derivation to be able to use the Fourier representation of the Gaussian process proposed by \citet{VFF}. This allows a choice of kernels in the \Matern family. Additionally, we show how the posterior process can be approximated using standard MCMC methods \cite{SparseMCMC}. This allows flexible representations of the posterior over functions $f(\cdot)$, and a Bayesian treatment of the hyperparameters, whereas Lloyd et al's approach was restricted to a Gaussian approximation for $f(\cdot)$ and point estimates of the hyperparameters.

Second, we deliver insights and improvements to the model. The use of a square-root link function can result in \emph{nodal lines}: where the latent function $f(\cdot)$ crosses zero, the rate function $\lambda(\cdot)$ must approach zero on both sides, resulting in regions of the posterior that have unsatisfactory artifacts. In addition, these nodal lines create multiple modes in the posterior that are hard to sample and difficult to approximate well with a Gaussian approach. We investigate these effects and demonstrate how they can be mitigated by suitable specification of the prior parameters.

Third, motivated by spatiotemporal problems where a periodic function is expected \textit{a priori}, we construct a periodic kernel that can be combined with the Fourier features approach to \Matern kernels under our inference scheme.
We also show how to obtain a closed-form confidence interval for the rate $\lambda$ under a Gaussian approximation.

The key result of our work is that we can now scale Bayesian point process inference on a single GPU to hundreds of thousands of events from multiple observations with complex spatiotemporal patterns. We demonstrate this on the Porto taxi data set \cite{PortoData}.

In short, the contributions of this paper are 1) extending the model to include a constant mean function that mitigates the effect of nodal lines; 2) derivations to enable \Matern family kernels, their sums and products to be combined with variational Fourier features, both for variational inference and MCMC; 3) constructing parametric periodic kernels that can be used in this framework; 4) showing how we predict uncertainty in the inferred rate; 5) applying this to large-scale models with complex three-dimensional patterns.

\section{Cox process inference}

A Cox process is an inhomogeneous Poisson process, where the rate $\lambda$ is itself a stochastic process, hence also called ``doubly stochastic Poisson process''.
%
%
The probability of the data $\data = \{ \vec{x}_n \}_{n=1}^N$ under an inhomogeneous Poisson process with known rate function $\lambda(\cdot)$ is given by
\begin{equation}
  p(\data \given \lambda(\cdot)) = \exp\!\Big(\!-\! \int_\domain \lambda(\vec{x}) \dd{\vec{x}} \Big) \prod\limits_{n} \frac{\lambda(\vec{x}_n)^{n_{x_n}}}{n_{x_n}!} ,
  \label{eq:poisson}
\end{equation}
where \domain is the domain of the observation and $n_{x_n}$ is the multiplicity of events at location $\vec{x}_n$. In the following, we assume that all events are distinct and $n_{x_n} = 1$. 

We want to infer the posterior distribution of the rate function given an observation $\data$, which is given by
\begin{equation*}
	p(\lambda(\cdot) \given \data) = \frac{ p(\lambda(\cdot)) p(\data \given \lambda(\cdot)) }{
	\int p(\lambda(\cdot)) p(\data \given \lambda(\cdot)) \dd{\lambda(\cdot)} } .
\end{equation*}
The marginal likelihood in the denominator involves a double integral, which
gives rise to the so-called ``double intractability'' \cite{MurrayThesis}.

\subsection{Multiple observations}

We can also consider multiple observations corresponding to several draws from the same distribution. For example, when considering the rate of taxi pickups in a city, we may have separate observations $\data_o = \{\vec{x}^{(o)}_{n_o}\}_{n_o=1}^{N_o}$ from several days, $o=1,\dots,\Nobs$, each containing $N_o$ events. Instead of modelling each day independently, we may want to model the pickup rate of the average day, while still making use of all available data.
Assuming that observations are i.i.d.\ with the same rate function $\lambda(\cdot)$,
the likelihood is the product of \cref{eq:poisson} for each observation:
\begin{equation}
	p(\data_1, \dots, \data_\Nobs \given \lambda(\cdot)) = \mysmallprod\nolimits_{o=1}^\Nobs p(\data_o \given \lambda(\cdot)) .
\end{equation}

\subsection{Cox process constructions using Gaussian processes}

We model the intensity function $\lambda(\cdot)$ using a Gaussian process, $f(\cdot) \drawnfrom \GP$.
Whereas $f(\cdot)$ is unbounded between $-\infty$ and $+\infty$, the rate $\lambda(\cdot)$ needs to be non-negative everywhere. This is achieved by an inverse link function $\rho(\cdot)$ that ensures $\lambda(\cdot) = \rho(f(\cdot)) \ge 0$ for any value of $f$.
\citet{Lloyd2015} used $\lambda(\cdot)=f(\cdot)^2$, which has the advantage of tractable analytic derivations and linear scaling in the number of data.

However, the lack of injectivity of $f^2$ can lead to nodal lines that will be discussed in more detail in \cref{sec:nodal}.
To mitigate this disadvantage, we extend the link function to include a constant offset $\beta$, so that $\lambda(\cdot) = (f(\cdot) + \beta)^2$.
Our model is
\begin{equation}
	p(\data, f(\cdot), \hypers) = p(\data \given \lambda = (f(\cdot)+\beta)^2 ) \allowbreak p(f(\cdot) \given \hypers) \allowbreak p(\hypers) ,
\end{equation}
where $\hypers$ contains $\beta$ and the kernel hyperparameters.
We make this tractable by approximating the exact GP posterior $p(f(\cdot), \hypers \given \data)$ by $q(f(\cdot)\given\hypers) q(\hypers)$, where $q(f(\cdot) \given \hypers) = \GP(f; \tilde{\mu}(\cdot), \tilde{\Sigma}(\cdot,\cdot))$. In expectation, the inferred rate is then $\expectation_{q(f(\cdot))} \lambda(\cdot) = \tilde{\mu}(\cdot)^2 + \tilde{\Sigma}(\cdot)$. We now derive the inference objective, and later discuss our choices for $q$.

\subsection{Objective}

Due to our extended model specification, we rederive the Evidence Lower Bound Objective (ELBO) from \citet{Lloyd2015}.
We optimize $q(f(\cdot))$ by minimizing the KL divergence to the true posterior, where we leave the dependence on the hyperparameters implicit:\footnote{The full derivation is given in the Supplementary Material.}
\begin{align}
	\mathcal{K} &= \KL{q(f(\cdot))}{p(f(\cdot) \given \data)} \nonumber \\
	&= - \expectation_{q(f(\cdot))} \Big[\! \log \frac{p(f(\cdot), \data)/p(\data)}{q(f(\cdot))} \Big] \nonumber \\
	&= -\expectation_q \Big[\! \log \frac{p(f(\cdot))}{q(f(\cdot))} \Big] - {\color{likelihood} \expectation_q \big[\! \log p(\data \given f(\cdot)) \big] } + \log p(\data) \nonumber \\
	&= \KL{q(f(\cdot))}{p(f(\cdot))} - {\color{likelihood} \mathcal{L}_D } + \log p(\data) \nonumber \\
	&= \log p(\data) - \mathcal{L} ,
\end{align}
where $\mathcal{L}$ is the ELBO from \citet{Lloyd2015}.
Minimizing $\mathcal{K}$ with respect to a variational distribution $q(f(\cdot))$ is equivalent to maximizing the ELBO $\mathcal{L}$.
We use the result from \citet{KLbound} that the KL between the approximate posterior and prior processes is a KL divergence at the inducing points.
For details, see \citet{MatthewsThesis}.

For our Cox process model, the likelihood term ${\color{likelihood} \mathcal{L}_D }$ is
\begin{align}
	{\color{likelihood} \mathcal{L}_D } &= \expectation_{q(f(\cdot))} [ \log p(\{\data_o\}_{o=1}^\Nobs\} \given f(\cdot)) ] \nonumber \\
	&= - \Nobs \, {\color{areaterm} \expectation_{q(f(\cdot))} \Big[ \int_\domain \lambda(\vec{x}) \dd\vec{x} \Big] } \nonumber \\
	&\quad + {\color{dataterm} \sum\nolimits_{o=1}^\Nobs \sum\nolimits_{n_o=1}^{N_o} \expectation_{q(f(\cdot))} [ \log \lambda(\vec{x}^{(o)}_{n_o}) ] } \nonumber \\
	&= - \Nobs {\color{areaterm} \mathcal{L}_{f_x} } + {\color{dataterm} \mathcal{L}_{f_n} } ,
\end{align}
where we allow for multiple observations, and the sum in ${\color{dataterm} \mathcal{L}_{f_n} }$ extends over the $N = \sum_{o=1}^\Nobs N_o$ events in all observations.
We now consider these terms separately for our choice of link function, $\lambda(\vec{x}) = ( f(\vec{x}) + \beta )^2$.

The data term ${\color{dataterm} \mathcal{L}_{f_n} }$ aims to increase $\lambda$ at the event locations:
\begin{equation}
	{\color{dataterm} \mathcal{L}_{f_n} } = \sum\nolimits_{n=1}^N \expectation_{q(f(\cdot))} [\log [( f(\vec{x}_n) + \beta )^2]] .
	\label{eq:lfn-term}
\end{equation}
Each summand in the ${\color{dataterm} \mathcal{L}_{f_n} }$ term corresponds to an integral
\begin{equation*}
	\int_{-\infty}^\infty \log[(f_n + \beta)^2] \, \normaldist(f_n; \tilde{\mu}(\vec{x}_n), \tilde{\Sigma}(\vec{x}_n,\vec{x}_n) \dd{f_n} ,
\end{equation*}
where $f_n=f(\vec{x}_n)$.
With a change of variables $f = f_n + \beta$, this becomes equivalent to the one-dimensional integral that
\citet{Lloyd2015} showed can be evaluated in closed form:
\[
	\int_{-\infty}^\infty \log (f^2) \, \normaldist(f; \tilde{\mu}, \tilde{\sigma}^2) \dd{f}
	= - \tilde{G}\big(\!- \frac{\tilde{\mu}^2}{2 \tilde{\sigma}^2} \big) + \log\! \big(\frac{\tilde{\sigma}^2}{2} \big) - C
,\]
where $\tilde{G}(\cdot)$ can be represented by a lookup table.

The area term ${\color{areaterm} \mathcal{L}_{f_x} }$ aims to minimize the overall rate, which ensures that $\lambda$ becomes small where there are fewer events:
\begin{align}
	{\color{areaterm} \mathcal{L}_{f_x} }& = \expectation_{q(f(\cdot))} \Big[ \int_\domain ( f(\vec{x}) + \beta )^2 \dd\vec{x} \Big] \nonumber \\
	& = \!\! \int_\domain \!\!\expectation_q [f(\vec{x})^2] \dd\vec{x} + 2 \beta \!\! \int_\domain \!\!\expectation_q [f(\vec{x})] \dd\vec{x} + \beta^2 |\domain| .
\end{align}
These integrals over the moments of $f(\cdot)$ are similar to the ``$\Psi$ statistics'' that show up in GPs with uncertain input \cite{GPLVM}, but over a uniform rather than Gaussian distribution.
They depend on the form of $q(f(\cdot))$, which we discuss in the following for a Gaussian variational approximation and variationally sparse MCMC.

\section{Approximations}

\subsection{Variational Gaussian approximation}
To keep the GP tractable, we use a sparse approximation to the full posterior.
We consider a set of inducing points $\mat{Z} = \{\vec{z}_m\}_{m=1}^M$, and we collect the inducing variables $u_m = f(\vec{z}_m)$ in a vector $\uu = f(\mat{Z})$. Our approximation matches the prior conditioned on the values at the inducing points: $q(f(\cdot) \given \uu) = p(f(\cdot) \given \uu)$.
We define $\ku(\cdot) = \cov(\uu, f(\cdot)) = k(\mat{Z}, \cdot)$ and $\Kuu = \cov(\uu, \uu) = k(\mat{Z}, \mat{Z})$.


Assuming a Gaussian form for the variational approximate distribution, $q(\uu) = \normaldist(\vec{m}, \mat{S})$,
then
\begin{equation}
    \begin{split}
	q(f(\cdot)) &= \int p(f(\cdot) \given \uu) q(\uu) \dd{\uu} = \GP(f; \tilde{\mu}(\cdot), \tilde{\Sigma}(\cdot, \cdot)) , \quad\text{where} \\
	\tilde{\mu}(\cdot) &= \ku(\cdot)\transposed \Kuu^{-1} \vec{m} , \\
	\tilde{\Sigma}(\cdot,\cdot) &= k(\cdot,\cdot) - \ku(\cdot)\transposed ( \Kuu^{-1} - \Kuu^{-1} \mat{S} \Kuu^{-1} ) \ku(\cdot) .
    \end{split}
    \label{eq:vb:meancov}
\end{equation}
Similar to the derivation by \citet{Lloyd2015}, we have
\begin{align}
	{\color{areaterm} \mathcal{L}_{f_x} } &=
	\vec{m}\transposed \Kuu^{-1} \mat{\Psi} \Kuu^{-1} \vec{m}
	+ \sigma^2 \area{\domain}
	- \tr (\Kuu^{-1} \mat{\Psi}) \nonumber \\
	&\quad + \tr(\Kuu^{-1} \mat{S} \Kuu^{-1} \mat{\Psi})
	+ 2\beta \vec{\Phi}\transposed \Kuu^{-1} \vec{m} + \beta^2 \area{\domain} ,
	\label{eq:lfx-variational}
\end{align}
where $\sigma^2 = k(x,x)$ is the variance of the kernel (assuming stationarity), $\mat{\Psi} = \int_\domain \ku(\vec{x}) \ku(\vec{x})\transposed \dd{\vec{x}}$, and $\vec{\Phi} = \int_\domain \ku(\vec{x}) \dd{\vec{x}}$.
Note that the terms involving $\beta$ in \cref{eq:lfx-variational} are missing in \citet{Lloyd2015}.
The expectation integrals $\mat{\Psi}$ and $\vec{\Phi}$ for Fourier features will be derived in \cref{sec:vffpsi}.
We can now optimize $\mathcal{L}$ with respect to $\vec{m}$, $\mat{S}$, and $\hypers$, which takes \order{NM^2} operations per gradient step.

\subsection{Sparse MCMC}
Though the variational Gaussian approximation often works well in practice, in some cases it may be important to determine the posterior distribution of the hyperparameters instead of using point estimates, and not restrict the form of $q(f(\cdot))$.
\citet{SparseMCMC} introduced MCMC for variationally sparse GPs. This approach still relies on a sparse set of inducing features to describe the function, but does not restrict $q(\uu)$ to be Gaussian.
They demonstrate by rearranging the objective $\mathcal{K}$ that the optimal variational distribution is
\begin{equation}
	\log \hat{q}(\uu, \hypers) = {\color{likelihood} \mathcal{L}_D' } + \log p(\uu \given \hypers) + \log p(\hypers) - \log Z ,
\end{equation}
where ${\color{likelihood} \mathcal{L}_D' } = \expectation_{p(f(\cdot) \given \uu, \hypers)} [\log p(\data \given f(\cdot))]$,
and the constant $Z$ normalizes the distribution.
We can sample from this distribution using MCMC. Here, we use Hamiltonian MC.

In each step, we need to evaluate $\log \hat{q}(\uu, \hypers)$ up to a constant: $\log \pi(\uu, \hypers) = {\color{likelihood} \mathcal{L}_D' } + \log p(\uu\given\hypers) + \log p(\hypers)$,
where the expectation in ${\color{likelihood} \mathcal{L}_D' }$ is under the conditional distribution
\begin{equation}
	p(f(\cdot) \given \uu, \hypers) = \GP(f; \ku(\cdot)\transposed \Kuu^{-1} \uu, \allowbreak k(\cdot,\cdot) - \ku(\cdot)\transposed \Kuu^{-1} \ku(\cdot)) .
	\label{eq:mcmc:fvconditional}
\end{equation}
For our Cox process model, the ${\color{areaterm} \mathcal{L}'_{f_x} }$ term is now a function of $\uu$, as opposed to $\vec{m}$ and $\mat{S}$ in \cref{eq:lfx-variational}:
\begin{equation}
    {\color{areaterm} \mathcal{L}'_{f_x} } =
	\uu\transposed \Kuu^{-1} \mat{\Psi} \Kuu^{-1} \uu
	+ \sigma^2 \area{\domain}
	- \tr(\Kuu^{-1} \mat{\Psi})
	+ 2\beta \vec{\Phi}\transposed \Kuu^{-1} \uu
	+ \beta^2 \area{\domain} .
\end{equation}
The ${\color{dataterm} \mathcal{L}'_{f_n} }$ term is similar to \cref{eq:lfn-term}, with the mean and variance of the Gaussian distribution now given by \cref{eq:mcmc:fvconditional}.
This MCMC approach requires \order{NM} computations in each step for the evaluation of $\log \pi(\vec{u}, \hypers)$.

\subsection{Fourier features}
\label{sec:vff}

Point process models need to accurately describe the rate function across the entire domain, not just where events are occurring. For this reason, it is ben\-e\-fi\-cial to use an approximation that has support everywhere, instead of at a fixed number of inducing points.

When data are contained in a smaller submanifold of the domain, inducing points are an effective way of approximating the GP.
However, for modelling point processes we need to be able to describe the entire domain, as the rate depends on both mean and variance of the underlying GP.
Not observing any events in a region is also informative (in that it suggests a lower rate): we need inducing points everywhere in the domain, even where there are no observations.

The Fourier features introduced by \citet{VFF} have support across the entire domain.
They are based on a spectral representation of the GP, related to random Fourier features \cite{RandomFourierFeatures}, but with fixed frequencies.
The random variables corresponding to these features are given by\footnote{Where $\langle g(\cdot), h(\cdot) \rangle_{\mathcal{H}}$ represents the inner product in the Reproducing Kernel Hilbert Space $\mathcal{H}$ associated with the kernel $k(\cdot,\cdot)$.} $u_m = \langle \phi_m(\cdot), f(\cdot) \rangle_{\mathcal{H}}$, where
\begin{multline}
	\vec{\phi}(x) = [1, \cos(\omega_1(x-a)), \dots, \cos(\omega_M(x-a)), \\
	\sin(\omega_1(x-a)), \dots, \sin(\omega_M(x-a))]\transposed .
\end{multline}
The features are parametrized by the frequencies (we choose $\omega_m = 2\pi m/(b-a)$ for $m=1,\dots,M$) and the bounding box $[a,b]$. The bounding box is part of the approximation and should be chosen somewhat larger than the domain \domain.

It can be proved that this choice of $\uu$ leads to $\ku(\cdot) = \vec{\phi}(\cdot)$.
Note that the features do not depend on the kernel parameters. This allows us to calculate the $\mat{\Psi}$ matrix very efficiently, as it does not change within each iteration of the optimization.
The covariance between features, $\cov(u_m, u_{m'}) = \langle \phi_m, \phi_{m'} \rangle_{\mathcal{H}}$, turns out to have low rank:
$
	\Kuu = \diag{\vec{\alpha}} + \mat{W} \mat{W}\transposed 
$,
where the rank of $\mat{W}$ depends on the kernel (one for \Matern-$\sfrac{1}{2}$, two for \Matern-$\sfrac{3}{2}$, and three for \Matern-$\sfrac{5}{2}$).
This means that we can evaluate $\Kuu^{-1} \ku(\cdot)$ in \order{N M} operations rather than \order{N M^2}.

The Fourier features can only be applied to one-dimensional kernels. However, we can apply this framework to higher-dimensional domains by using additive (sum) and/or separable (product) kernels.
For a sum kernel,
$
	k(\vec{x},\vec{x}')=\sum_{d=1}^D k_d(x_d, x_d')
$,
the feature matrix is given by the stacking of the one-dimensional feature matrices. Features in different dimensions do not interact, and \Kuu is block-diagonal.
For a product kernel,
$
	k(\vec{x}, \vec{x}') = \prod_{d=1}^D k_d(x_d, x_d')
$,
the features are given by
$
  \vec{\phi}(\vec{x}) = \kron_{d=1}^D [\phi_1(x_d), \dots, \phi_{M}(x_d)]\transposed
$.
Again, this is independent of the kernel parameters.
\Kuu has Kronecker structure:
$
  \cov(\uu, \uu) = \Kuu = \kron_{d=1}^D \Kuu^d
$.
We follow \citet{VFF}'s recommendation in using a sum-of-Kronecker structure for $\mat{S}$.

%

\subsection{$\Psi$ statistics for Fourier features}
\label{sec:vffpsi}

To implement Fourier features for the Cox process model discussed in this paper, we also need $\mat{\Psi}$ and $\vec{\Phi}$, which we derive in this section.

For Fourier features, $\ku(\cdot)$ only consists of cosine and sine functions, independent of the kernel, so $\mat{\Psi}$ and $\vec{\Phi}$ are just integrals over (products of) cosine functions.
As they do not depend on any hyperparameters, we can simply precompute $\ku(\cdot)$, $\mat{\Psi}$, and $\vec{\Phi}$ outside the optimization loop.
In 1D, with $\domain = [c,d]$, we derive (\cf Supplementary Material)
\newcommand*{\co}[2]{\mathcal{C}_{#1}^{#2}}
\newcommand*{\so}[2]{\mathcal{S}_{#1}^{#2}}
\begin{equation}
	\mat{\Psi} = \!\! \int_c^d \ku(x) \ku(x)\transposed \dd{x}
	= \begin{bmatrix}
		\area{\domain}        & \Psi_{(1,i)}            & \Psi_{(1,j)} \\
		\Psi_{(1,i)}\transposed & \Psi_{(i,i)}            & \Psi_{(i,j)} \\
		\Psi_{(1,j)}\transposed & \Psi_{(i,j)}\transposed & \Psi_{(j,j)}
	\end{bmatrix} \!\!,
\end{equation}
with
$\co{m}{x} = \cos(\omega_m x)$ and $\so{m}{x} = \sin(\omega_m x)$ and
\begin{align*}
	\Psi^{(1,i)}_m &= \omega_m^{-1} [\so{m}{d} - \so{m}{c}] , \\
	\Psi^{(1,j)}_m &= -\omega_m^{-1} [\co{m}{d} - \co{m}{c}] , \\
	\Psi^{(i,i)}_{m=n} &= \area{\domain}/2 + \omega_{4m}^{-1} (\so{2m}{d} - \so{2m}{c}) , \\
	\Psi^{(i,i)}_{m\neq n} &= \frac{n (\co{m}{d}\so{n}{d} - \co{m}{c}\so{n}{c}) - m (\so{m}{d}\co{n}{d} - \so{m}{c}\co{n}{c})} {2\pi(n^2-m^2) / (b - a)} , \\
	\Psi^{(j,j)}_{m=n} &= \area{\domain}/2 - \omega_{4m}^{-1} (\so{2m}{d} - \so{2m}{c}) , \\
	\Psi^{(j,j)}_{m\neq n} &= \frac{m (\co{m}{d}\so{n}{d} - \co{m}{c}\so{n}{c}) - n (\so{m}{d}\co{n}{c} - \so{m}{c}\co{n}{c})} {2\pi(n^2-m^2) / (b - a)} , \\
	\Psi^{(i,j)}_{m=n} &= \omega_{4m}^{-1} [\co{2m}{d} - \co{2m}{c}] , \\
	\Psi^{(i,j)}_{m\neq n} &= \frac{n (\so{m}{d}\so{n}{d} - \so{m}{c}\so{n}{c}) + m (\co{m}{d}\co{n}{d} - \co{m}{c}\co{n}{c})} {2\pi(n^2-m^2) / (b - a)} .
\end{align*}
%
For multi-dimensional sum and product kernels, the $\mat{\Psi}$ matrix can be constructed from the 1D cases. For a product kernel, $\mat{\Psi}$ is the Kronecker product of the 1D $\mat{\Psi}$ matrices for each dimension. For a sum kernel, $\mat{\Psi}$ is block-diagonal; the diagonal blocks are equivalent to the 1D $\mat{\Psi}$ matrices scaled by the total volume divided by the length of that dimension. The off-diagonal blocks are the outer product of the first rows of the 1D $\mat{\Psi}$ matrices scaled by the total volume divided by the lengths of the two involved dimensions.

The first Fourier feature is the constant $1$, so in 1D the $\vec{\Phi}$ vector is equal to the first row of $\mat{\Psi}$. For multiple dimensions using a product kernel, $\vec{\Phi}$ is the Kronecker-vector-stack of the first rows of the component matrices of $\mat{\Psi}$. For a sum kernel, $\vec{\Phi}$ in each dimension is equivalent to the corresponding 1D case, scaled by the total volume divided by the length of that dimension.


\section{Methods}

\subsection{Parametric periodic kernels}
\label{sec:periodic}

When we have more explicit prior knowledge about the behavior of our data, we can encode this in parametric kernels. For example, in spatiotemporal point processes, we may want to model periodic components such as time-of-day, where we want an explicitly periodic behavior on the time dimension to avoid discontinuities at the ``roll-over point''.

Given a vector of features $\vec{\phi}(\cdot)$, we can define a parametric kernel as $k(\cdot, \cdot) = \vec{\phi}(\cdot)\transposed \mat{G}^{-1} \vec{\phi}(\cdot)$ using a positive-definite Gram matrix $\mat{G}$. With $u_m = \langle \phi_m(\cdot), f(\cdot)\rangle_{\mathcal{H}}$, this results in $\Kuu = \mat{G}$ and $\ku(\cdot)=\vec{\phi}(\cdot)$.\footnote{This is similar to the VFF method, except that the process is completely determined by \uu, in the sense that the conditional variance $f \given \uu$ is zero.}

To obtain a periodic kernel, we can use sines and cosines with the right periodicity, \ie $\sin(\omega_m \cdot)$ and $\cos(\omega_m \cdot)$ with $\omega_m = 2\pi m / T$, where $m \in \mathbb{N}$ and $T = d - c$ is the period. Note that these features fit neatly into the Fourier feature framework, which means that the calculation of $\mat{\Psi}$ and $\vec{\Phi}$ remains the same as before (now with $[a, b] = [c, d]$).

If we choose a matrix $\mat{G}$ with non-zero off-diagonal elements, the correlations between different basis functions lead to a dependence on the start of the domain (\ie the results are no longer invariant under translation of the data along the periodic dimension).
We avoid this by setting $\mat{G}=\diag{s(\omega_m)^{-1}}$, where $s(\omega)$ is the spectral density of a \Matern kernel. This results in the same kernel as the \textit{sparse spectrum GP} \cite{quinonero2010sparse}, but whilst periodicity was avoided in that work by randomizing the frequencies, we have deliberately selected the frequencies to capture a suitable prior over periodic functions.

There is a non-identifiability issue between variance and lengthscale: increasing the lengthscale reduces the overall mass $\sum_{m=1}^M s(\omega_m)$, and mimics a reduction in variance. This makes it hard to optimize or do inference on the hyperparameters.
This issue can be avoided by normalizing the discrete spectrum and only introducing the variance $\sigma^2$ at the end as a multiplicative scaling factor:
$
	s'(\omega_i) = \sigma^2 s(\omega_i) / \big(\! \sum_{m=1}^M s(\omega_m) \!\big) 
$.

\subsection{Uncertainty prediction in variational inference}

Beyond inferring a mean intensity function, we can also compute an uncertainty interval for the intensity function, given the data.

Due to the highly skewed likelihood of the Poisson process, it does not make sense to consider the variance of $\lambda$ directly.
However, we can evaluate the percentiles numerically. We have $f(\vec{x}^*) \drawnfrom \normaldist(\tilde{\mu}, \tilde{\sigma}^2)$ (\cf \cref{eq:vb:meancov}). First, we normalize,
$g = f / \tilde{\sigma}$, so that
$
	g \drawnfrom \normaldist(\tilde{\mu}/\tilde{\sigma}, 1)
$. Then $g^2 = f^2 / \tilde{\sigma}^2$ is distributed according to a non-central $\chi^2$ distribution,
$
	g^2 \drawnfrom \chi^2(k=1, \lambda=\tilde{\mu}^2/\tilde{\sigma}^2)
$,
for which there exist standard library functions to evaluate the percentiles.
We can backtransform these by $f^2 = g^2 \tilde{\sigma}^2$. We evaluate this using the conditional mean and variance of the posterior GP at each point for which we want to compute the percentiles.

\subsection{Test set likelihood}

To compare models and evaluate their performance, we need to calculate the test set likelihood. This is difficult in a point process model due to the double intractability.
The probability density for a test set $\testdata = \{\vec{x}^{*}_n\}_{n=1}^{N^{*}}$ given our model trained on the training set $\data$ is exactly given by
\begin{equation}
	p(\testdata \given \data) = \mysmallint p(\testdata \given f(\cdot)) p(f(\cdot) \given \data) \dd{f(\cdot)} ,
\end{equation}
where $p(\testdata \given f(\cdot)) = \exp(- \int_\domain \lambda(\vec{x}) \dd{\vec{x}} ) \prod_{n=1}^{N^{*}} \lambda(\vec{x}^{*}_n)$ (assuming $\Nobs=1$) with $\lambda(\cdot) = (f(\cdot) + \beta)^2$.
We generally want to calculate log densities:
\begin{equation}
	\testllh{exact} = \log p(\testdata \given \data)
	\approx \log \!\mysmallint\! q(f(\cdot)) p(\testdata \given f(\cdot)) \dd{f(\cdot)} ,
	\label{eq:logllh-exact}
\end{equation}
where we approximated the true posterior by $q(f)$.
\citet{Lloyd2015} further approximate this using Jensen's inequality, similar to the ELBO:
\begin{equation}
	\testllh{ELBO} = \mysmallint q(f(\cdot)) \log p(\testdata \given f(\cdot)) \dd{f(\cdot)} \le \testllh{exact} .
\end{equation}
However, there is no guarantee that $\testllh{ELBO} \approx \testllh{exact}$.
Instead, we consider the mean likelihood that assumes an inhomogeneous Poisson process with a deterministic rate function that is given by the mean $\hat{\lambda}(\cdot) = \tilde{\mu}(\cdot)^2 + \tilde{\sigma}^2(\cdot)$:
\begin{equation}
	\testllh{mean} = \log p(\testdata \given \hat{\lambda}) = - \! \int_\domain \! \hat{\lambda}(\vec{x}) \dd{\vec{x}} + \! \mysmallsum\limits_{n=1}^{N^{*}} \hat{\lambda}(\vec{x}^{*}_n) .
	\label{eq:logllh-mean}
\end{equation}
We can approximate the integral in the exact test set log density by drawing samples from the posterior, $f_s \drawnfrom q(f(\cdot))$:
\begin{equation}
	\testllh{exact} \approx \testllh{sample} = \log \mfrac{1}{N_s} \mysmallsum\nolimits_{s=1}^{N_s} p(\testdata \given f_s(\cdot)) .
	\label{eq:logllh-sample}
\end{equation}
In practice, the samples need to be evaluated on a fine grid, which is infeasible in multiple dimensions.
We show for the 1D case that \cref{eq:logllh-mean} and \cref{eq:logllh-sample} generally correlate well (\cf \cref{fig:llhcomparison} in the Supplementary Material). If not specified otherwise, in this paper we use \cref{eq:logllh-mean} to calculate test set likelihoods.

\subsection{Initialization}
\label{sec:init}

When optimizing non-convex objective functions, a reasonable starting value for the unknown variables can help to avoid bad local optima. Here we discuss our approaches to initializing hyperparameters and variational distribution.

\myparagraph{Constant offset.}
We can estimate $\beta$ from the events count. This is the correct answer if the rate is completely uniform: $\bar{\lambda} = \average{N_o} / \area{\domain}$, where $\average{N_o} = \sum_{o=1}^{\Nobs} N_o / \Nobs$.
We then initialize $\beta = \sqrt{\bar{\lambda}}$.
This is generally an overestimate; due to the square-root link function, positive $f$ will influence the mean more strongly than negative $f$. We can account for this empirically by adjusting $\beta$ downwards, \eg $\beta = 2/3 \bar{\lambda}^{1/2}$.

\myparagraph{Kernel lengthscale.}
We can obtain a good starting point for the lengthscale hyperparameters by visually inspecting the spatial distribution of data or using the estimated bandwidth from kernel intensity smoothing (KIS).

\myparagraph{Kernel variance.}
For a ``wiggly'' rate, the magnitude scale of the GP $f(\cdot)$ should be on the order of $\beta$ to be able to describe regions of low intensity. In this case, we can initialize the variance $\sigma^2 \approx \beta^2 \approx \bar{\lambda}$.
Again, this is generally an overestimate, and for more even distributions of events the variance will be lower. This can be determined through visual inspection, based on KIS or otherwise.

For MCMC, we put Gamma priors on the hyperparameters, with shape and scale chosen such that mean of the prior corresponds to the point estimate discussed before, and the standard deviation of the prior is sufficiently smaller than the mean. This gives the MCMC the freedom to explore but ensures that there is zero mass at zero.

\myparagraph{Variational approximate distribution.}
In variational inference, we also need to initialize the mean $\vec{m}$ and covariance $\mat{S}$ of the approximating Gaussian distribution. When using an appropriate initial value for $\beta$, in practice, it works well to simply initialize with a zero mean, $\vec{m} = 0$.
To obtain different starting conditions, we can draw $\vec{m} \drawnfrom p(\uu) = \normaldist(\vec{0}, \Kuu)$.

In order for the KL term in the ELBO not to swamp the other terms (which would make optimization difficult), it is advantageous to initialize with the covariance matrix of the prior, $\mat{S} = \Kuu$.
When using product kernels with Fourier features in two or more dimensions, we want $\mat{S}$ to be a sum of two Kronecker-structured matrices; in this case it is not possible to simply set $\mat{S} = \Kuu$, but we can approximate it.
Note that the two summands need to be different to break the symmetry.
Ensuring that the summands are not related by a simple scale factor is crucial to avoid duplicate eigenvalues.
We choose $\mat{S}^i_1 = 0.2 \Kuu^i$ and $\mat{S}^i_2 = \Kuu^i + 0.2$.

\subsection{Nodal lines}
\label{sec:nodal}

\begin{figure}
	\centering
	\setlength{\figurewidth}{\columnwidth}
	\setlength{\figureheight}{0.35\columnwidth}
	\pgfplotsset{group/vertical sep=5mm}
	\pgfplotsset{axis x line=middle}
	\pgfplotsset{axis y line=left}
	\pgfplotsset{every inner x axis line/.append style={-}}
	\input{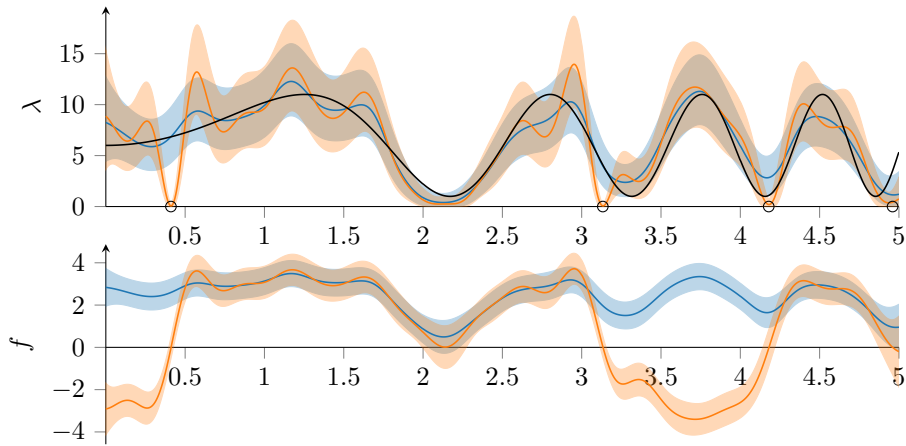}
	\caption{A close-up of the result for $\lambda_2$ and $\Nobs=10$ in \cref{fig:1d}, showing the effect of nodal lines in a model fit (orange) and a better model fit without zero crossings (blue). Where $f$ crosses zero, $\lambda$ also collapses to zero (black circles). This is even the case at a point where the actual rate is high. The nodal lines lead to a short lengthscale estimation and oscillations throughout the domain.}
    \label{fig:nodal}
\end{figure}

Since the inverse link function $\lambda(\cdot)=f(\cdot)^2$ is not injective, different $f(\cdot)$s can lead to similar rates.
For example, $\pm f(\cdot)$ have the same associated rate.
A pathological case can arise when a rate is low but not zero around some location $x_0$ of the input space.
Consider a 1D example, in which a function $f_1(\cdot)$ is positive in some regions and negative in others, with a zero crossing at $x_0$. This may be a good initial guess early on in the optimization, especially if there are regions with low intensity, and, hence, no direct penalty at the zero crossing.
However, this leads to a nodal line\footnote{We refer to these artifacts as \textit{nodal lines} because of their resemblance to that effect in wave interference.} at $x_0$, where the rate is forced to be zero.
A better candidate $f_2(\cdot)$ would be positive everywhere and low around $x_0$. 
Both $f_1$ and $f_2$ constitute local optima, but stepping from one to the other is almost impossible because of the high energy barrier that separates the two (half of $f_1(\cdot)$ would need to change sign, and the intermediate steps would be highly penalized). 
Once a zero crossing exists, it is unlikely to ever disappear again, which severely restricts the optimization.
One example is shown in \cref{fig:nodal}.
This is particularly an issue in higher dimensions, and when regions with large numbers of events alternate with low-event regions.
When using MCMC, this leads to many local modes, with no mixing between them. For the chains to be able to explore, we need to end up in the mode with $f(\cdot)$ positive everywhere.

We mitigate this problem by including the constant offset term $\beta$, as $(f + \beta)$ is less likely to cross zero. 
It is also important to choose a sufficiently large number of inducing features (frequencies or points). If there are not enough inducing features to represent small-scale variations, we are more likely to end up with nodal lines, as they make it easier for the intensity to quickly go to zero and back up again.

\section{Empirical results}

\begin{figure}[t]
	\centering
	\setlength{\figurewidth}{0.4\textwidth}
	\setlength{\figureheight}{0.35\textwidth}
	\pgfplotsset{group/horizontal sep=2mm,group/vertical sep=2mm}
	\pgfplotsset{every axis title/.append style={at={(0.5,0.8)}}}
	\pgfplotsset{axis x line=middle}
	\pgfplotsset{axis y line=left}
	\pgfplotsset{every inner x axis line/.append style={-}}
	\input{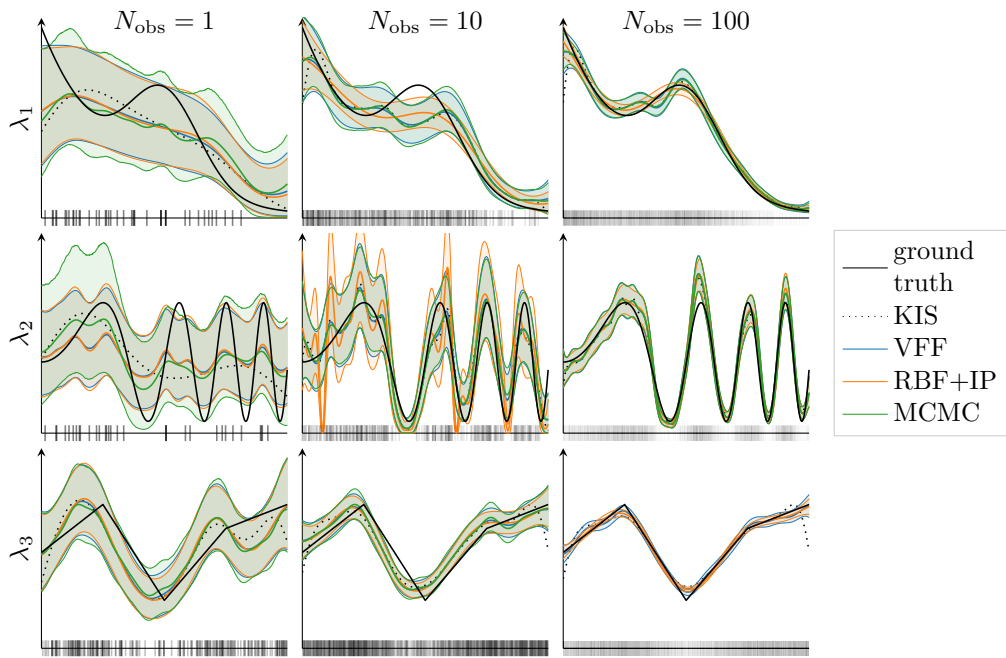}
	\caption{Inferred rate function for the 1D systems $\lambda_1$, $\lambda_2$, and $\lambda_3$ from \citet{AdamsMurrayMacKay} (rows from top to bottom), based on 1, 10, or 100 draws from the distribution (columns from left to right). We compare the optimized results for VFF (blue) and RBF+IP (orange) with KIS (dotted) and VFF+MCMC (green). VFF and RBF+IP give comparable results, and we show the 5/95 percentile interval. With an increased number of observations, the inferred mean gets closer to the ground truth (black), and the confidence interval becomes tighter. Models are based on $M=40$ frequencies and $M=40$ inducing points.}
    \label{fig:1d}
\end{figure}

We compare variational Fourier features (VFF) with the inducing point approach using the radial basis function (RBF) kernel\footnote{Also known as squared exponential, Gaussian, or exponentiated quadratic kernel.} (denoted RBF+IP).
For VFF we use the \Matern-$\sfrac{5}{2}$ kernel, as its behavior is similar to the RBF kernel.
As the inducing features need to provide support across the entire domain, we put inducing points on a regular $D$-dimensional grid, without optimizing their positions.
As a baseline, we include KIS with edge correction \cite{KIS}, optimized using the leave-one-out objective.
For MCMC we make use of whitening, representing $\uu = \mat{R} \vec{v}$, where $\mat{R} \mat{R}\transposed = \Kuu$ and $\vec{v} \drawnfrom \normaldist(\vec{0}, \mat{I})$.

\subsection{1D comparison}

We use synthetic 1D examples to show that Fourier features have the same expressive power as RBF+IP and to demonstrate our methods: uncertainty prediction, MCMC, and sampled vs.\ mean test set likelihood.
We use the three example intensity functions from \citet{AdamsMurrayMacKay} as shown in Figure \ref{fig:1d} and described in the Supplementary Material. 

For variational inference, hyperparameters are initialized as in \cref{sec:init}.
We train VFF, RBF+IP, and KIS on training sets containing 1, 10, and 100 draws from the ground truth. The mean inferred intensity and the 5/95 percentiles are shown in \cref{fig:1d}. We calculate test set likelihoods both using $\mathcal{L}^{*}_\text{sample}$ and $\mathcal{L}^{*}_\text{mean}$. \Cref{fig:llhcomparison} in the Supplementary Material shows that they correlate well with each other.
For each, the likelihoods for VFF and RBF+IP differ by less than 1\%, demonstrating that both approaches have the same capability of modelling these simple data.
The resulting point estimates for the hyperparameters are compared with the initial values in \cref{fig:hyper} in the Supplementary Material.
For MCMC, the posterior distributions for the hyperparameters are shown in \cref{fig:mcmchyper} in the Supplementary Material.
For 22400 data points and 20 frequencies, MCMC takes about \SI{3.5}{\second} per sample; it scales linearly in $N$ and $M$.

\subsection{Real world data: Porto taxi pickups}

The real advantage of VFF is in modelling complex, large-scale data in higher dimensions.
Here we apply our model to real-world data from the Porto taxi trajectory data set \cite{PortoData}.
This contains \num{1.7e6} trajectories covering the entire year from 1 July 2013 to 30 June 2014. We selected only the pickup locations of the trajectories to model.
We left the taxi ranks aside (these can be modeled separately, when needed) and focus on the pickups scattered throughout the downtown area.\footnote{We manually removed the area around Cais da Estiva and Cais da Ribeira from the analysis; this has a very sharp peak due to the density of bars, and is more appropriately modeled as another ``taxi rank''.} This results in a data set with 462k pickups across 365 days.
We focus on two separate regimes, a spatial model trained on a single day and a spatiotemporal model trained on 100 days.

In each case, we consider multiple optimizations from different initial values for kernel variance, $\sigma^2 \in \{\bar{\lambda}, 1/2 \bar{\lambda}\}$, and constant offset, $\beta \in \{\bar{\lambda}^{1/2}, 2/3\bar{\lambda}^{1/2}\}$.
We initialize the mean of the variational appromixate distribution both to $\vec{m}=0$ and to random draws $\vec{m}\drawnfrom\normaldist(0,\Kuu)$.

\begin{figure}
	\centering
	\setlength{\figurewidth}{5.65cm}
	\setlength{\figureheight}{3.5cm}
	\resizebox{\textwidth}{!}{%
	\pgfplotsset{group/horizontal sep=2mm,group/vertical sep=8mm}
	\pgfplotsset{every axis title/.append style={at={(0.5,0.85)}}}
	\input{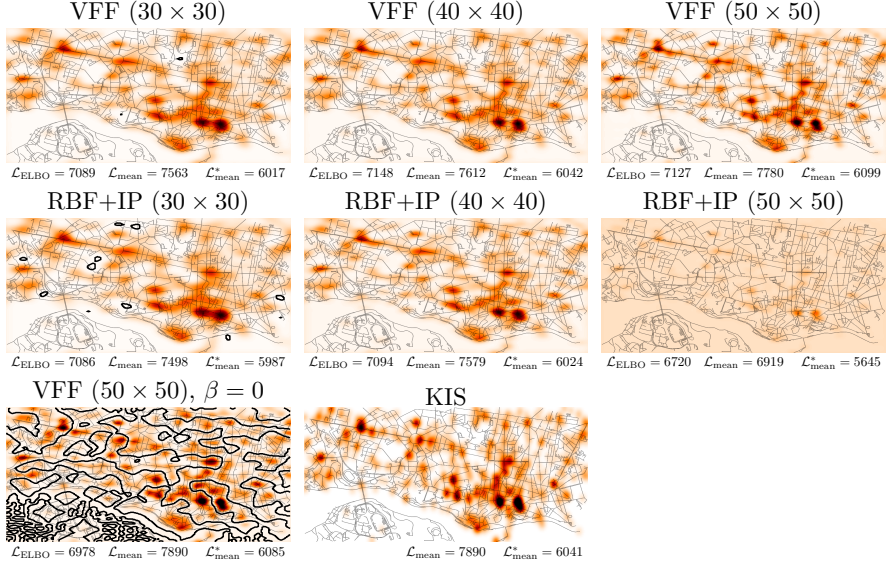}
	}
	\caption{Inferred rate function for a 2D model trained on one day of the Porto taxi data set. We compare KIS, VFF and RBF+IP for different numbers of inducing frequencies/points (all figures have the same color scale). We give ELBO and the mean likelihoods for training set, $\mathcal{L}_\text{mean}$, and test set, $\mathcal{L}^{*}_\text{mean}$. Including more frequencies leads to a better-resolved rate function, and nodal lines (black contours) disappear. RBF+IP takes a long time and is very hard to optimize; after \num{48} core-hours the $50\times50$ model still shows a very flat rate function. At the example of VFF with $50\times50$ frequencies, we show how fixing $\beta=0$ leads to many nodal lines. The KIS model overfits.}
    \label{fig:2d}
\end{figure}

\paragraph{2D comparison.}
Here we show the benefit of Bayesian inference when the data are sparse. The training data are the 1089 pickups in the downtown area from Monday 1 July 2013.
We compare variational inference using VFF and RBF+IP and KIS.
\Cref{fig:2d} shows that by using a product kernel and VFF we can resolve features better than RBF+IP. Moreover, optimization was much faster. VFF took about one second per gradient step, resulting in run times on a single core of \SIrange{15}{30}{\minute} for $30\times30$ frequencies, \SI{1}{\hour} for $40\times40$, and \SIrange{2}{4}{\hour} for $50\times50$.
RBF+IP took \SIrange{50}{500}{\second} per gradient step, depending on the size of the grid, and the optimization run times were \SI{12}{\hour} ($30\times30$ inducing points), \SI{39}{\hour} ($40\times40$), and for $50\times50$ inducing points, the optimization still had not converged after \SI{48}{\hour}.
Moreover, we found that for the RBF+IP model, many optimization runs failed due to numerical instabilities.
Without the constant offset (fixing $\beta=0$), the model ends up with nodal lines. This is shown for one example fit in \cref{fig:2d}.
%
%
%
%

The model fit with the highest ELBO does not necessarily describe the best fit overall.
For example, for the $50\times50$ VFF model, the fit with the highest ELBO had larger lengthscales and both lower training and lower test set likelihoods than other fits of the same model with a slightly lower ELBO.
For the RBF model, we found a model fit that has a very low ELBO, but very high likelihood (both training and test set), though with a severe amount of nodal lines.
For different hyperparameters, the gap (given by the KL term) may be different. This results in a bias towards a different lengthscale.
For this reason, we chose model fits manually from the different runs based on crossvalidation -- comparing the training set likelihoods and inferred mean intensity for those model fits that had the highest ELBO values.
This indicates that point estimates of the hyperparameters are a bad idea. We need good priors to find the local minimum that best describes the data.

\paragraph{Large-scale demonstration.}

Here we demonstrate that our method can scale to very large data sets using variational inference. We consider a 3D spatiotemporal model where we include time-of-day. As the training set we chose the first 100 odd weekdays.
This results in a data set with \num{113 020} events. 

We chose $35\times35$ frequencies for the spatial dimensions, and a periodic kernel based on the \Matern-$\sfrac{5}{2}$ spectrum as described in \cref{sec:periodic} with \num{25} frequencies for the time dimension.
Optimizing the model for \num{20000} gradient steps took ca.\ \SIrange{24}{28}{\hour} on a Tesla P100 GPU.
We are mainly limited by the amount of GPU memory; the memory requirements grow linearly with the number of data points (and linearly with the number of frequencies in any one dimension, or cubically with the number of frequencies for all three dimensions).

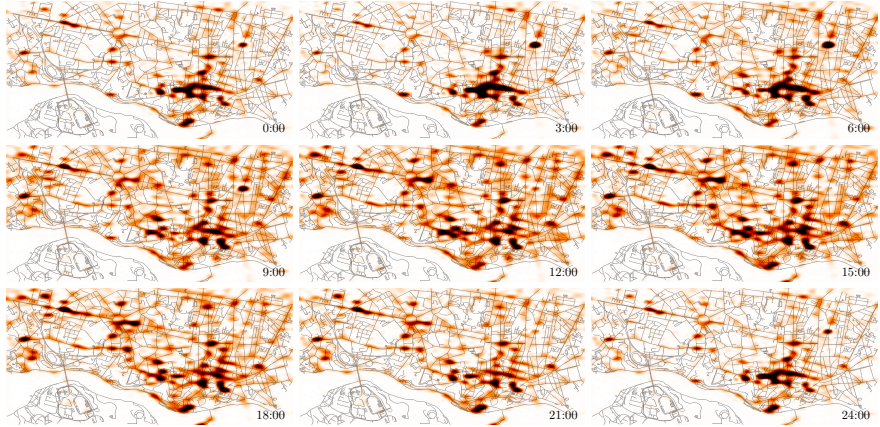
\begin{figure}[t]
	\centering
	\resizebox{\textwidth}{!}{%
	\setlength{\figurewidth}{5.6cm}
	\setlength{\figureheight}{3.5cm}
	\pgfplotsset{group/horizontal sep=1mm,group/vertical sep=1mm}
	\input{Fig_porto_3d.tikz}
	}
	\caption{Time slices through the 3D spatiotemporal model of the rate of taxi pickups, trained on 113k events from the Porto taxi data set. A video is available as part of the Supplementary Material.}
    \label{fig:3d}
\end{figure}

In \cref{fig:3d} we show time slices through the resulting 3D model at every three hours.
Note that it shows improved spatial resolution compared to the 2D model.
Without the periodic kernel on the time dimension, the inferred rate for 00:00 would not match up with that for 24:00, see \cref{fig:notperiodic} in the Supplementary Material.

We model the rate of events with large spatial and temporal variation across a large city area. Our model is continuous in space and time, does not rely on a pre-determined grid, and we can encode the periodicity inherent in the temporal domain. To the best of our knowledge, this is the first time a principled Bayesian model of this scale has been made possible.
%

\section{Discussion and Conclusions}

We have presented two variational approaches to inference in Gaussian process modulated Poisson processes, using Gaussian approximations and MCMC.  We have shown that our Fourier-based approximations are effective and robust, and scale to huge datasets.

We have discussed the effect of \emph{nodal lines} in the model which are due to the square-root link function: we showed how to mitigate these effects by selecting suitable hyperparameter initial conditions (for optimization) and priors (for MCMC). We found empirically that the choice of optimizer and initial condition can have a huge effect on the resulting model, and discussed the use of the ELBO for selecting between different local optima.

\clearpage
\bibliography{references}
\bibliographystyle{icml2018}

\clearpage
\section*{Supplementary Material}

\setcounter{figure}{1}
\renewcommand\thefigure{S\arabic{figure}}

\appendix

\section{Example rate functions}
The three example functions in Figure \ref{fig:1d} are
\begin{align}
	\lambda_1(s) &= 2 \exp(-s/15) + \exp(-((s-25)/10)^2) , \\
	\lambda_2(s) &= 5 \sin(s^2) + 6 , \\
	\lambda_3(s) &= \text{piecewise linear} ,
\end{align}
where $\lambda_3$ goes through the following points: $(0, 2), (25, 3), (50, 1), (75, 2.5), (100, 3)$.
The domains are $\domain_1 = [0, 50]$, $\domain_2 = [0, 5]$, and $\domain_3 = [0, 100]$.
The average number of events per draw are $\bar{\lambda}_1 = 46.92$, $\bar{\lambda}_2 = 33.49$, and $\bar{\lambda}_3 = 224.37$.

\section{Derivation of objective}

We arrive at the our objective by considering the KL divergence between the true posterior $p(f(\cdot) \given \data)$ and our sparse approximation $q(f(\cdot))$:
\begin{align}
	p(f^* \given y) &= \int\int p(f^* \given f, \hypers) p(f, \hypers \given \data) \dd{\hypers}\dd{f} \\
	q(f^*) &= \int\int p(f^* \given u, \hypers) q(u, \hypers) \dd{\hypers}\dd{f}
\end{align}
We want to minimize their KL divergence, writing out the full probability distribution of everything:
\begin{align}
	\mathcal{K} &= \KL{q(f^*, f, u, \hypers)}{p(f^*, f, u, \hypers \given \data)} \\
	&= - \expectation_{q(f^*, f, u, \hypers)} \big[ \log \frac{p(f^* \given u, f, \hypers) p(u \given f, \hypers) p(f, \hypers \given \data)}{p(f^* \given u, f, \hypers) p(f \given u, \hypers) q(u, \hypers)} \big] \\
	&\overset{(a)}{=} - \expectation_{q(f, u, \hypers)} \big[ \log \frac{p(u \given f, \hypers) p(\data \given f, \hypers) p(f \given \hypers) p(\hypers) / p(\data)}{p(f \given u, \hypers) q(u, \hypers)} \big] \\
	&= - \expectation_{q(f, u, \hypers)} \big[ \log \frac{p(u \given f, \hypers) p(f \given \hypers) p(\data \given f, \hypers) p(\hypers) / p(\data)}{p(f \given u, \hypers) q(u, \hypers)} \big] \\
	&\overset{(b)}{=} - \expectation_{q(f, u, \hypers)} \big[ \log \frac{p(f \given u, \hypers) p(u \given \hypers) p(\data \given f, \hypers) p(\hypers) / p(\data)}{p(f \given u, \hypers) q(u, \hypers)} \big] \\
	&= - \expectation_{q(f, u, \hypers)} \big[ \log \frac{p(u \given \hypers) p(\data \given f, \hypers) p(\hypers) / p(\data)}{q(u, \hypers)} \big] \\
	&= - \expectation_{q(f, u, \hypers)} \big[ \log \frac{p(u \given \hypers) p(\data \given f, \hypers) p(\hypers)}{q(u, \hypers)} \big] + \log p(\data)
\end{align}
where we made use of (a)
\begin{equation}
	p(f, \hypers \given \data) = p(f, \hypers, \data) / p(\data) = p(\data \given f, \hypers) p(f \given \hypers) p(\hypers) / p(\data)
\end{equation}
and (b)
\begin{equation}
	p(u \given f, \hypers) p(f \given \hypers) = p(u, f \given \hypers) = p(f \given u, \hypers) p(u \given \hypers)
\end{equation}
\begin{align}
	\mathcal{K} &= -\expectation_{q(f,u,\hypers)} \log \frac{p(u \given \hypers)p(\hypers)}{q(u, \hypers)} - \expectation_{q(f,u,\hypers)} \log p(\data \given f, \hypers) + \log p(\data) \\
	&= -\expectation_{q(u,\hypers)} \log \frac{p(u,\hypers)}{q(u,\hypers)} - \expectation_{q(f,\hypers)} \log p(\data \given f, \hypers) + \log p(\data) \\
	&= \KL{q(u, \hypers)}{p(u, \hypers)} - \mathcal{L}_D + \log p(\data) \\
	&= \log p(\data) - \mathcal{L}
\end{align}
So with respect to a variational distribution $q(u)$, maximizing the ELBO $\mathcal{L}$ is equivalent to minimizing the KL divergence $\mathcal{K}$.


\section{Derivation of $\Psi$ matrix for Fourier features}
\label{sec:psi}

We want to calculate the matrix
\begin{equation}
	\mat{\Psi} = \int_\domain \ku(\vec{x})\transposed \ku(\vec{x}) \dd{\vec{x}} ,
\end{equation}
where $\ku(\cdot) = \vec{\phi}(\cdot)$.
We first calculate the elements of $\mat{\Psi}$ for a one-dimensional kernel, $\Psi_{ij} = \int \phi_i(x) \phi_j(x) \dd{x}$.

\newcommand*{\cosM}{\cos_{\omega_m}}
\newcommand*{\cosN}{\cos_{\omega_n}}
\newcommand*{\sinM}{\sin_{\omega_m}}
\newcommand*{\sinN}{\sin_{\omega_n}}

\subsection{Notation}

We use the following short-hand notation:
\begin{equation}
    \cosM = \cos \big( \omega_m ( x - a ) \big) , \qquad\text{where }\omega_m = \frac{2 \pi m}{b - a}
\end{equation}
where $m$ is an integer and $\omega_m$ is the corresponding natural frequency on the interval $[a,b]$.
and equivalently $\cosN$, $\sinM$, and $\sinN$. The Fourier features can then be written as
\begin{align}
    \phi_i (x) =
    \begin{cases}
        \cos_{\omega_i}     & \text{for } 0 \le i \le M \\
        \sin_{\omega_{i-M}} & \text{for } M < i \le 2 M
    \end{cases}
\end{align}

\subsection{One-dimensional kernel}

\newcommand*{\mynextcase}[1]{\paragraph*{#1}}

For Fourier features with $M$ frequencies, the first element in the feature vector is the constant $1$, then there are $M$ cosine functions, then $M$ sine functions.
This leads to six types of integrals:
$1\times1$, $1\times\cos$, $1\times\sin$, $\cos\times\cos$, $\cos\times\sin$, $\sin\times\sin$,
and we have to distinguish between features with the same frequency ($m=n$) or different frequencies ($m\neq n$).

For a domain $\domain = [c, d]$ in one dimension, we need to evaluate the integrals
\begin{equation}
	\Psi_{i,j} = \int_c^d \phi_i(x) \phi_j(x) \dd{x} .
\end{equation}
This can be split into the following cases:

\mynextcase{$\phi_0 \phi_0 = 1 \times 1$}

\begin{equation}
    \Psi_{0,0} = \int_c^d 1 dx = d - c
\end{equation}

\mynextcase{$\phi_0 \phi_i [1 \le i \le M] = 1 \times \cosM$, $m \ge 1$}

$1 \le i \le M$:
\begin{align}
    \Psi_{0,i} &= \int_c^d \cosM dx = \frac{b-a}{2\pi m} \sinM |_c^d \\
    &= \frac{b-a}{2\pi m} \big( \sin(\omega_m (d - a)) - \sin(\omega_m (c - a)) \big)
\end{align}

\mynextcase{$\phi_0 \phi_i [M < i \le 2 M] = 1 \times \sinM$, $m \ge 1$}

$M < i \le 2 M$:
\begin{align}
    \Psi_{0,i} &= \int_c^d \sinM dx = - \frac{b-a}{2\pi m} \cosM |_c^d \\
    &= - \frac{b-a}{2\pi m} \big( \cos(\omega_m (d - a)) - \cos(\omega_m (c - a)) \big)
\end{align}

\mynextcase{$\phi_i \phi_i [M < i \le 2 M] = \sinM \sinM$, $m = n$}

$M < i \le 2 M$:
\begin{align}
    \Psi_{i,i} &= \int_c^d \sinM \sinM dx = \int_c^d \sinM^2 dx \\
    &= \int_c^d \frac{1}{2} (1 - \cos_{\omega_{2m}}) \\
    &= \frac{1}{2} (d - c) - \frac{b-a}{2 \times 2\pi \times 2m} \sin_{\omega_{2m}} |_c^d \\
    &= \frac{1}{2} (d - c) - \frac{b-a}{8\pi m} \sin_{\omega_{2m}} |_c^d
\end{align}

\mynextcase{$\phi_i \phi_j [M < i, j \le 2 M] = \sinM \sinN$, $m \neq n$}

$M < i, j \le 2 M$:
\begin{align}
    \Psi_{i,j} &= \int_c^d \underbrace{ \sinM }_{u} \underbrace{ \sinN }_{v'} dx \\
    &\qquad | u' = \frac{2\pi m}{b-a}\cosM, \qquad v = -\frac{b-a}{2\pi n}\cosN \\
    &= - \frac{b-a}{2\pi n} \sinM \cosN |_c^d + \frac{m}{n} \int_c^d \underbrace{ \cosM }_{u} \underbrace{ \cosN }_{v'} dx \\
    &\qquad | u' = \frac{2\pi m}{b-a} (-\sinM), \qquad v = -\frac{b-a}{2\pi n}\sinN \\
    &= - \frac{b-a}{2\pi n} \sinM \cosN |_c^d + \frac{m}{n} \bigg[ \frac{b-a}{2\pi n} \cosM \sinN |_c^d + \frac{m}{n} \int_c^d \sinM \sinN dx \bigg] \\
    (1 - \frac{m^2}{n^2}) \int \dots &= - \frac{b-a}{2\pi n} \sinM \cosN |_c^d + \frac{b-a}{2\pi n} \frac{m}{n} \cosM \sinN |_c^d \\
    \int \dots &= \frac{n^2}{n^2 - m^2} \frac{b-a}{2\pi} \big( - \frac{1}{n} \sinM \cosN |_c^d + \frac{m}{n^2} \cosM \sinN |_c^d \big) \\
    &= \frac{1}{n^2 - m^2} \frac{b - a}{2\pi} \big(m \cosM \sinN |_c^d - n \sinM \cosN |_c^d \big)
\end{align}

\mynextcase{$\phi_i \phi_j [1 \le j \le M, M < i \le 2 M] = \sinM \cosN$, $m \neq n$}

$1 \le j \le M, M < i \le 2 M$:
\begin{align}
    \Psi_{i,j} &= \int_c^d \underbrace{ \sinM }_{u} \underbrace{ \cosN }_{v'} dx \\
    &\qquad | u' = \frac{2\pi m}{b-a} \cosM , \qquad v = \frac{b-a}{2\pi n} \sinN \\
    &= \frac{b-a}{2\pi n} \sinM \sinN |_c^d - \frac{m}{n} \int_c^d \cosM \sinN dx
\end{align}
\begin{align}
    \int_c^d \underbrace{ \cosM }_{u} \underbrace{ \sinN }_{v'} & \\
    &\qquad | u' = - \frac{2\pi m}{b-a} \sinM , \qquad v = - \frac{b-a}{2\pi n} \cosN \\
    &= - \frac{b-a}{2\pi n} \cosM \cosN |_c^d - \frac{m}{n} \int_c^d \sinM \cosN dx
\end{align}
\begin{align}
    \Psi_{i,j} &= \frac{b - a}{2\pi n} \sinM \sinN |_c^d - \frac{m}{n} \bigg[ - \frac{b-a}{2\pi n} \cosM \cosN |_c^d - \frac{m}{n} \int_c^d \sinM \cosN dx \bigg] \\
    &= \frac{b-a}{2\pi n} \sinM \sinN |_c^d + \frac{m}{n} \frac{b-a}{2\pi n} \cosM \cosN |_c^d + \left(\frac{m}{n}\right)^2 \int_c^d \sinM \cosN dx \\
    (1 - (\frac{m}{n})^2) \int \dots &= \frac{b - a}{2\pi n} (\sinM \sinN + \frac{m}{n} \cosM \cosN)|_c^d \\
    \int \dots &= \frac{1}{1 - \frac{m^2}{n^2}} \frac{b-a}{2\pi n} \big( \sinM \sinN |_c^d + \frac{m}{n} \cosM \cosN |_c^d \big) \\
    &= \frac{1}{n^2 - m^2} \frac{b-a}{2\pi} \big( n \sinM \sinN |_c^d + m \cosM \cosN |_c^d \big)
\end{align}

\mynextcase{$\phi_i \phi_j [1 \le j \le M, i = j + M] = \sinM \cosM$, $m = n$}

$1 \le j \le M, i = j + M$:
\begin{align}
    \Psi_{i,i+M} &= \int_c^d \sinM \cosM dx = \frac{1}{2} \int_c^d \sin_{\omega_{2m}} dx \\
    &= \frac{1}{2} (- \frac{b-a}{2\pi \times 2m} \cos_{\omega_{2m}} |_c^d ) \\
    &= - \frac{b - a}{8\pi m} \cos_{\omega_{2m}} |_c^d
\end{align}

\mynextcase{$\phi_i \phi_i [1 \le i \le M] = \cosM \cosM$, $m = n$}

$1 \le i \le M$:
\begin{align}
    \Psi_{i,i} &= \int_c^d \cosM \cosM dx = \int_c^d \cosM^2 dx = \int_c^d (1 - \sinM^2) dx \\
    &= \int_c^d \frac{1}{2} (1 + \cos_{\omega_{2m}}) \\
    &= \frac{1}{2} (d - c) + \frac{b-a}{2 \times 2\pi \times 2m} \sin_{\omega_{2m}} |_c^d \\
    &= \frac{1}{2} (d - c) + \frac{b-a}{8\pi m} \sin_{\omega_{2m}} |_c^d
\end{align}

\mynextcase{$\phi_i \phi_j [1 \le i, j \le M] = \cosM \cosN$, $m \neq n$}

$1 \le i, j \le M$:
\begin{align}
    \Psi_{i,j} &= \int_c^d \underbrace{ \cosM }_{u} \underbrace{ \cosN }_{v'} \\
    &\qquad | u' = \frac{2\pi m}{b-a}(-\sinM), \qquad v = \frac{b - a}{2\pi n} \sinN \\
    &= \frac{b-a}{2\pi n} \cosM \sinN |_c^d + \frac{m}{n} \int_c^d \sinM \sinN dx \\
    &= \frac{b-a}{2\pi} \big( \frac{1}{n} \cosM \sinN |_c^d + \frac{m}{n} \frac{1}{n^2 - m^2} m \cosM \sinN |_c^d - \frac{m}{n} \frac{1}{n^2 - m^2} n \sinM \cosN |_c^d \big) \\
    &= \frac{b-a}{2\pi} \big( \cosM \sinN |_c^d \times \Big(\underbrace{ \frac{1}{n} + \frac{m}{n} \frac{1}{n^2 - m^2} }_{(n^2 - m^2 + m)/n}\Big) - \dots \big) \\
    &= \frac{b-a}{2\pi (n^2 - m^2)} \big( (n^2 - m^2) \frac{1}{n} \cosM \sinN + \frac{m}{n} m \cosM \sinN - \frac{m}{n} n \sinM \cosN \big)|_c^d \\
    &= \frac{b - a}{2\pi (n^2 - m^2)} \big(\cosM \sinN \big[\frac{n^2 - m^2}{n} + \frac{m^2}{n}\big] - m \sinM \cosN \big)|_c^d \\
    &= \frac{b - a}{2\pi (n^2 - m^2)} \big(n \cosM \sinN - m \sinM \cosN \big)|_c^d \\
\end{align}

\clearpage
\subsection{Sum kernel}

For a multi-dimensional sum kernel, the resulting $\Psi$ matrix has block structure.
The diagonal blocks $\Psi_{(i,i)}$ are equivalent to the one-dimensional case $\Psi^{(i)}$, except as we integrate over all dimensions, we get a factor $(d_j - c_j)$ for each dimension $j \neq i$. The off-diagonal blocks $\Psi_{(i,j)}$, $i\neq j$, correspond to the integrals
\[
    \int_{c_i}^{d_i} \phi_m(x_i) dx_i \int_{c_j}^{d_j} \phi_n(x_j) dx_j
\]
which are the outer product of the first rows of the corresponding one-dimensional $\Psi$ matrices, except again we get a factor $(d_k - c_k)$ for each dimension $k \notin {i,j}$. The calculation can be simplified by constructing diagonal blocks $\Psi^{(i)} / (d_i - c_i)$ and off-diagonal blocks
\[
    \Psi^{(i)}_{1,\cdot} \otimes \Psi^{(j)}_{1,\cdot} \frac{1}{(d_i - c_i) (d_j - c_j)}
\]
and finally scaling the overall matrix by the volume $V = \prod_i (d_i - c_i)$.

\subsection{Product kernel}

For a product kernel in $D$ dimensions, the different dimensions do not interact with each other, and the full $\Psi$ matrix is given by the Kronecker product of the one-dimensional matrices:
\begin{equation}
  \Psi = \bigkron_{d=1}^D \Psi_d .
\end{equation}

\clearpage
\section{Supplementary Figures}

\begin{figure}[h]
	\centering
	\resizebox{\textwidth}{!}{%
	\input{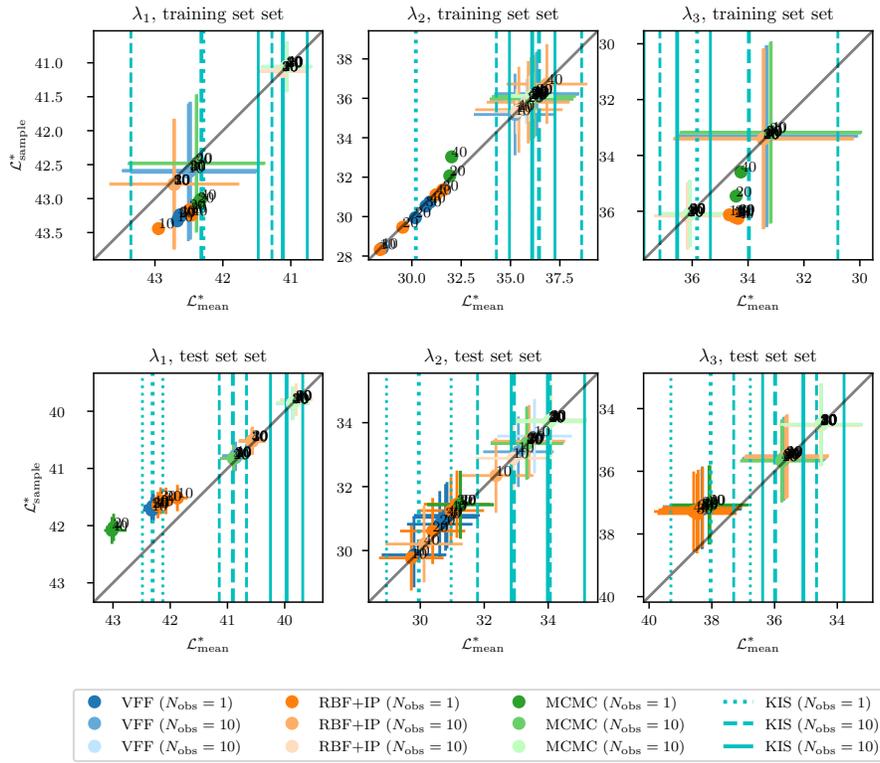}
	}
	\vspace{-1em}
	\caption{Test set likelihoods for the 1D examples $\lambda_1$, $\lambda_2$, and $\lambda_3$, comparing $\mathcal{L}^{*}_\text{sample}$ and $\mathcal{L}^{*}_\text{mean}$.
	We show results for VFF, RBF+IP, MCMC, and KIS, for different training set sizes and for different number of features (frequencies or points, numbers next to each dot).
	The test set contains $100$ observations; the error bars show the error of the mean across observations. For KIS, the confidence interval is denoted by the thinner lines.
	This shows that using the mean rate instead of the full samples is generally a good approximation. Exceptions to this are when the uncertainty in the posterior is large; this is more relevant for small numbers of observations in the training set.}
	\label{fig:llhcomparison}
\end{figure}

\begin{figure}
	\centering
	\resizebox{\textwidth}{!}{%
	\input{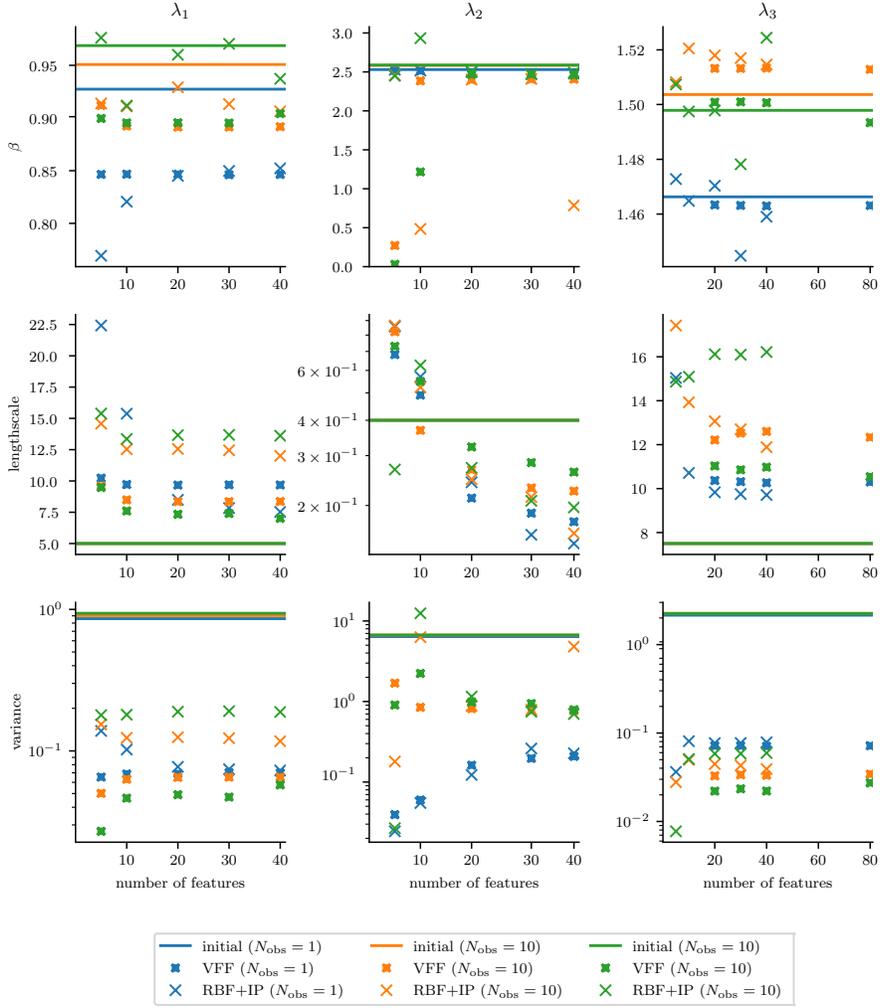}
	}
	\vspace{-2em}
	\caption{Optimized hyperparameters in variational inference for VFF and RBF+IP for the 1D examples $\lambda_1$, $\lambda_2$, and $\lambda_3$, for different numbers of observations $\Nobs$ in the training set. Horizontal lines denote the initial values. This shows how the point estimates converge with increasing number of features. VFF tends to converge faster. Note that an insufficient number of inducing features is generally associated with a too large lengthscale. The constant offset $\beta$ can be estimated well using our heuristics, whereas it is more difficult to estimate the variance \textit{a priori}.}
	\label{fig:hyper}
\end{figure}

\begin{figure}
	\centering
	\resizebox{\textwidth}{!}{%
	\input{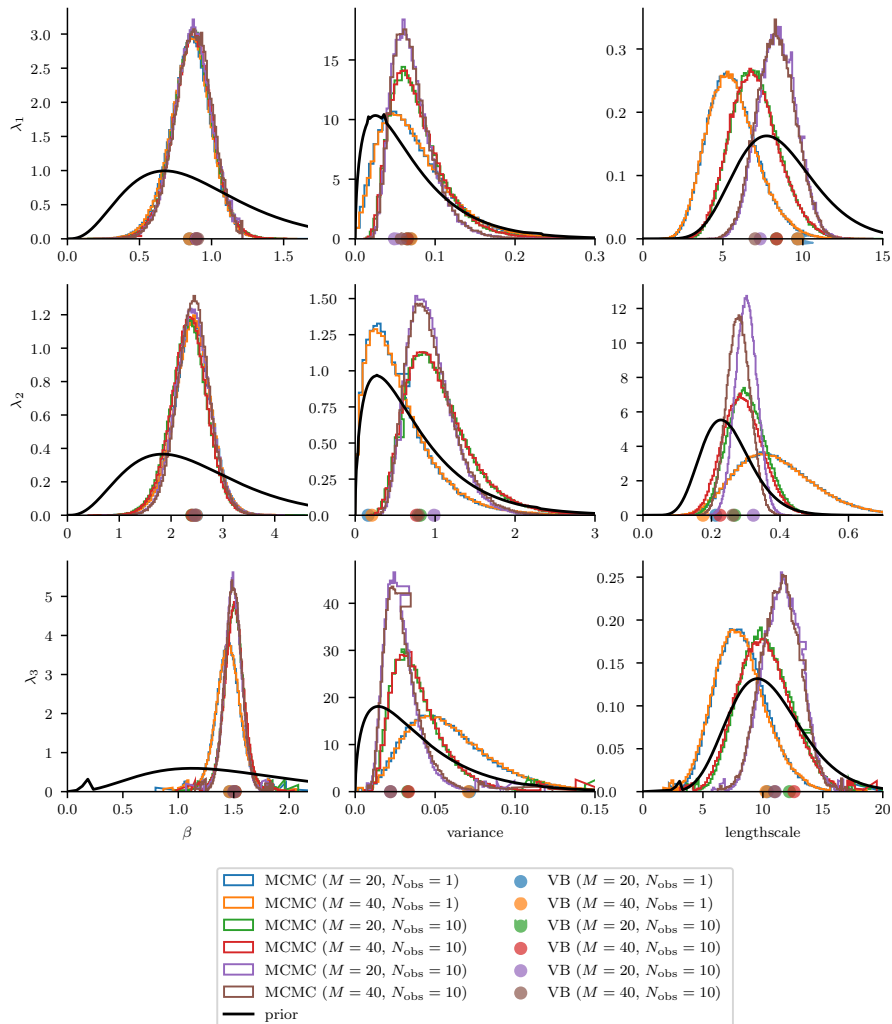}
	}
	\vspace{-1em}
	\caption{Histograms of the hyperparameter posterior distributions obtained through MCMC, for the same observations as in \cref{fig:hyper}. Dots denote corresponding VB point estimate. In most cases, $M=20$ and $M=40$ frequencies cannot be distinguished. The exception is $\lambda_2$ with $\Nobs=100$ observations, where a larger number of frequencies allows us to resolve the oscillations with a slightly smaller lengthscale. For the comparatively smooth $\lambda_1$ and $\lambda_3$ the lengthscale increases with the number of observations, as the data becomes more even and less affected by the shot noise. For $\lambda_2$, with less observations the inference smoothes over the oscillations. In all cases, larger numbers of observations result in tighter and more peaked distributions.}
	\label{fig:mcmchyper}
\end{figure}

\begin{figure}
	\centering
	\includegraphics[width=0.4\textwidth]{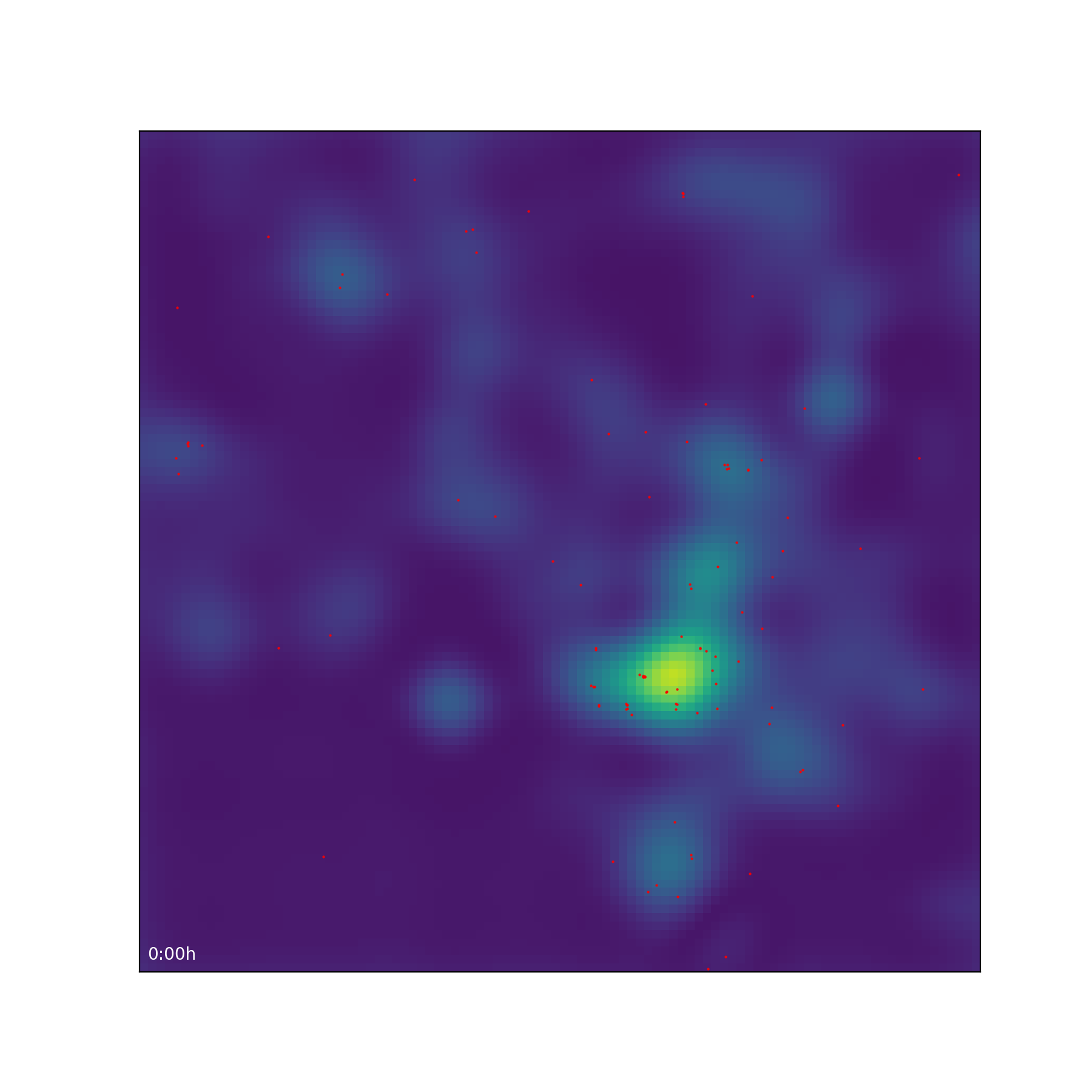}
	\includegraphics[width=0.4\textwidth]{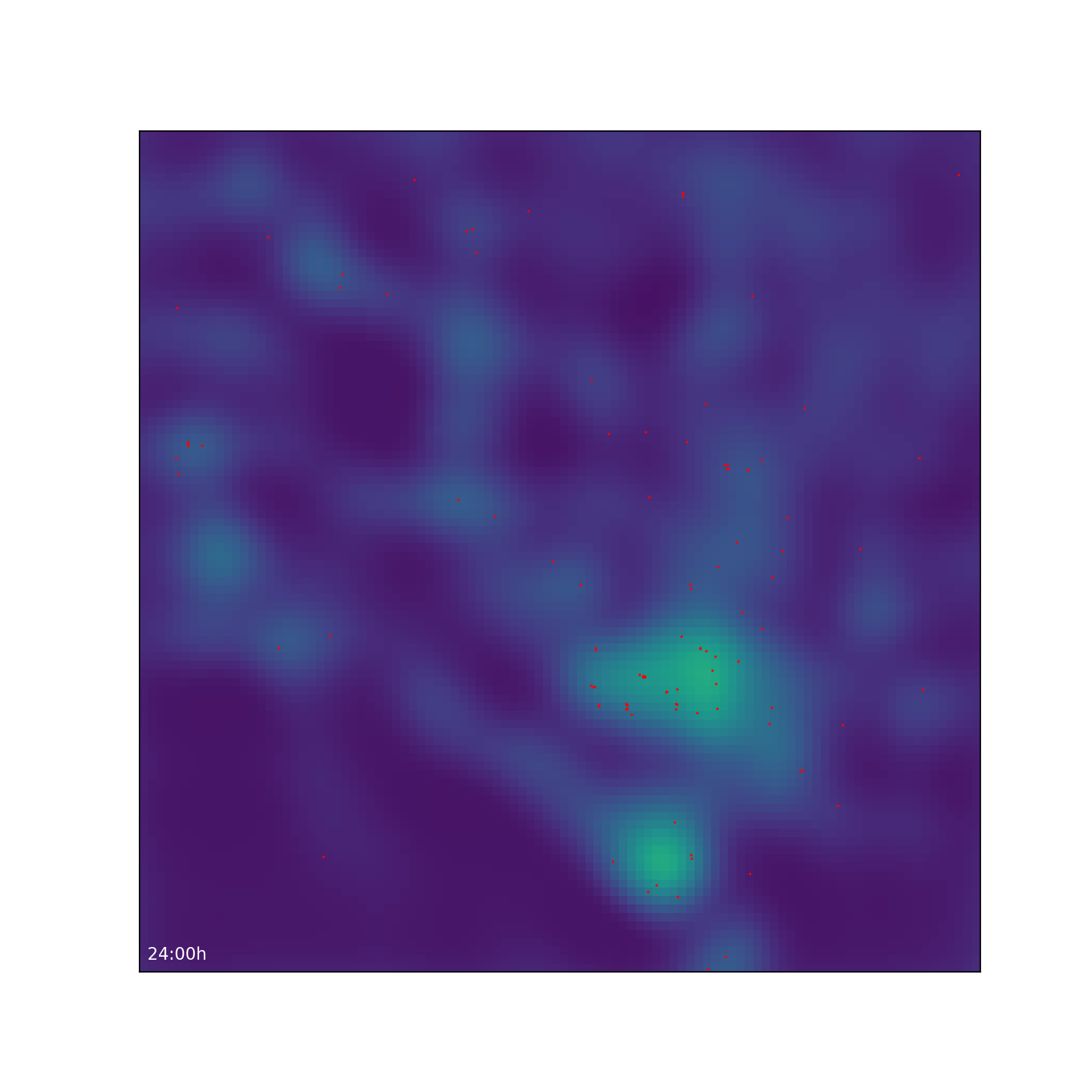}
	\caption{Inferred rate for midnight 00:00 (left) and 24:00 (right) for a spatiotemporal model including day-of-time but without a non-periodic kernel for the time dimension.
	With a periodic kernel, the rates for both cases would be equal.}
	\label{fig:notperiodic}
\end{figure}

\end{document}

%% file: setup.tex
\usepackage{graphicx}
\usepackage{booktabs}       
\usepackage{amsfonts}       
\usepackage{nicefrac}       
\usepackage{microtype}      
\usepackage{amsmath,amssymb,amsthm,mathrsfs,amsfonts,dsfont}
\usepackage{bm}
\usepackage{wrapfig}

\usepackage{hyperref}       
\usepackage{url}            
\usepackage{cleveref}
\usepackage{xspace}

%% file: abbrev.tex
\renewcommand*{\vec}[1]{\bm{\mathrm{#1}}}
\newcommand*{\mat}[1]{\mathrm{#1}}

\def\mysmallint{\begingroup\textstyle\int\endgroup}
\def\mysmallsum{\begingroup\textstyle\sum\endgroup}
\def\mysmallprod{\begingroup\textstyle\prod\endgroup}
\newcommand*{\order}[1]{\ensuremath{\mathcal{O}(#1)}}
\newcommand*{\data}{\ensuremath{\mathcal{D}}\xspace}
\newcommand*{\testdata}{\mathcal{D}^{*}}
\newcommand*{\testllh}[1]{\mathcal{L}^{*}_\text{#1}}
\newcommand*{\domain}{\ensuremath{\mathcal{T}}\xspace}
\newcommand*{\area}[1]{|#1|}
\newcommand*{\average}[1]{\langle#1\rangle}
\newcommand*{\normaldist}{\ensuremath{\mathcal{N}}}
\newcommand*{\drawnfrom}{\sim}
\newcommand*{\GP}{\ensuremath{\mathcal{GP}}}
\newcommand*{\given}{\,|\,}
\newcommand*{\hypers}{\ensuremath{\Theta}\xspace}
\newcommand*{\uu}{\vec{u}}
\newcommand*{\Kuu}{\ensuremath{\mat{K}_{\uu\uu}}\xspace}
\newcommand*{\ku}{\ensuremath{\vec{k}_{\vec{u}}}}

\newcommand*{\kron}{\otimes}
\newcommand*{\bigkron}{\bigotimes}
\newcommand*{\transposed}{^\intercal}
\newcommand*{\cov}{\operatorname{cov}}
\newcommand*{\diag}[2][]{\operatorname{diag}#1(#2#1)}
\newcommand*{\KL}[2]{\ensuremath{\operatorname{KL}[#1 \,\|\, #2]}}
\newcommand*{\expectation}{\mathbb{E}}
\newcommand*{\tr}{\operatorname{tr}}

\newcommand*{\dd}[1]{\,\mathrm{d}#1}

\newcommand*{\Matern}{Mat\'ern\xspace}
\newcommand*{\Nobs}{\ensuremath{{N_\text{obs}}}\xspace}

\newcommand*{\eg}{e.g., }
\newcommand*{\ie}{i.e., }
\newcommand*{\cf}{cf.\ }

%% file: Fig_porto_3d.tikz
\begin{tikzpicture}

\begin{groupplot}[group style={group size=3 by 3}]
\nextgroupplot[
hide x axis,
hide y axis,
xmin=0, xmax=1,
ymin=0, ymax=1,
width=\figurewidth,
height=\figureheight,
tick align=outside,
tick pos=left,
x grid style={white!69.019607843137251!black},
y grid style={white!69.019607843137251!black}
]
\addplot graphics [includegraphics cmd=\pgfimage,xmin=0, xmax=1, ymin=0, ymax=1] {Fig_porto_3d1.png};
\addplot graphics [includegraphics cmd=\pgfimage,xmin=0, xmax=1, ymin=0, ymax=1] {Fig_porto_3d2.png};
\node at (axis cs:0.99,0.01)[
  scale=0.5,
  anchor=south east,
  text=black,
  rotate=0.0
]{  0:00};
\nextgroupplot[
hide x axis,
hide y axis,
xmin=0, xmax=1,
ymin=0, ymax=1,
width=\figurewidth,
height=\figureheight,
tick align=outside,
tick pos=left,
x grid style={white!69.019607843137251!black},
y grid style={white!69.019607843137251!black}
]
\addplot graphics [includegraphics cmd=\pgfimage,xmin=0, xmax=1, ymin=0, ymax=1] {Fig_porto_3d3.png};
\addplot graphics [includegraphics cmd=\pgfimage,xmin=0, xmax=1, ymin=0, ymax=1] {Fig_porto_3d4.png};
\node at (axis cs:0.99,0.01)[
  scale=0.5,
  anchor=south east,
  text=black,
  rotate=0.0
]{  3:00};
\nextgroupplot[
hide x axis,
hide y axis,
xmin=0, xmax=1,
ymin=0, ymax=1,
width=\figurewidth,
height=\figureheight,
tick align=outside,
tick pos=left,
x grid style={white!69.019607843137251!black},
y grid style={white!69.019607843137251!black}
]
\addplot graphics [includegraphics cmd=\pgfimage,xmin=0, xmax=1, ymin=0, ymax=1] {Fig_porto_3d5.png};
\addplot graphics [includegraphics cmd=\pgfimage,xmin=0, xmax=1, ymin=0, ymax=1] {Fig_porto_3d6.png};
\node at (axis cs:0.99,0.01)[
  scale=0.5,
  anchor=south east,
  text=black,
  rotate=0.0
]{  6:00};
\nextgroupplot[
hide x axis,
hide y axis,
xmin=0, xmax=1,
ymin=0, ymax=1,
width=\figurewidth,
height=\figureheight,
tick align=outside,
tick pos=left,
x grid style={white!69.019607843137251!black},
y grid style={white!69.019607843137251!black}
]
\addplot graphics [includegraphics cmd=\pgfimage,xmin=0, xmax=1, ymin=0, ymax=1] {Fig_porto_3d7.png};
\addplot graphics [includegraphics cmd=\pgfimage,xmin=0, xmax=1, ymin=0, ymax=1] {Fig_porto_3d8.png};
\node at (axis cs:0.99,0.01)[
  scale=0.5,
  anchor=south east,
  text=black,
  rotate=0.0
]{  9:00};
\nextgroupplot[
hide x axis,
hide y axis,
xmin=0, xmax=1,
ymin=0, ymax=1,
width=\figurewidth,
height=\figureheight,
tick align=outside,
tick pos=left,
x grid style={white!69.019607843137251!black},
y grid style={white!69.019607843137251!black}
]
\addplot graphics [includegraphics cmd=\pgfimage,xmin=0, xmax=1, ymin=0, ymax=1] {Fig_porto_3d9.png};
\addplot graphics [includegraphics cmd=\pgfimage,xmin=0, xmax=1, ymin=0, ymax=1] {Fig_porto_3d10.png};
\node at (axis cs:0.99,0.01)[
  scale=0.5,
  anchor=south east,
  text=black,
  rotate=0.0
]{  12:00};
\nextgroupplot[
hide x axis,
hide y axis,
xmin=0, xmax=1,
ymin=0, ymax=1,
width=\figurewidth,
height=\figureheight,
tick align=outside,
tick pos=left,
x grid style={white!69.019607843137251!black},
y grid style={white!69.019607843137251!black}
]
\addplot graphics [includegraphics cmd=\pgfimage,xmin=0, xmax=1, ymin=0, ymax=1] {Fig_porto_3d11.png};
\addplot graphics [includegraphics cmd=\pgfimage,xmin=0, xmax=1, ymin=0, ymax=1] {Fig_porto_3d12.png};
\node at (axis cs:0.99,0.01)[
  scale=0.5,
  anchor=south east,
  text=black,
  rotate=0.0
]{  15:00};
\nextgroupplot[
hide x axis,
hide y axis,
xmin=0, xmax=1,
ymin=0, ymax=1,
width=\figurewidth,
height=\figureheight,
tick align=outside,
tick pos=left,
x grid style={white!69.019607843137251!black},
y grid style={white!69.019607843137251!black}
]
\addplot graphics [includegraphics cmd=\pgfimage,xmin=0, xmax=1, ymin=0, ymax=1] {Fig_porto_3d13.png};
\addplot graphics [includegraphics cmd=\pgfimage,xmin=0, xmax=1, ymin=0, ymax=1] {Fig_porto_3d14.png};
\node at (axis cs:0.99,0.01)[
  scale=0.5,
  anchor=south east,
  text=black,
  rotate=0.0
]{  18:00};
\nextgroupplot[
hide x axis,
hide y axis,
xmin=0, xmax=1,
ymin=0, ymax=1,
width=\figurewidth,
height=\figureheight,
tick align=outside,
tick pos=left,
x grid style={white!69.019607843137251!black},
y grid style={white!69.019607843137251!black}
]
\addplot graphics [includegraphics cmd=\pgfimage,xmin=0, xmax=1, ymin=0, ymax=1] {Fig_porto_3d15.png};
\addplot graphics [includegraphics cmd=\pgfimage,xmin=0, xmax=1, ymin=0, ymax=1] {Fig_porto_3d16.png};
\node at (axis cs:0.99,0.01)[
  scale=0.5,
  anchor=south east,
  text=black,
  rotate=0.0
]{  21:00};
\nextgroupplot[
hide x axis,
hide y axis,
xmin=0, xmax=1,
ymin=0, ymax=1,
width=\figurewidth,
height=\figureheight,
tick align=outside,
tick pos=left,
x grid style={white!69.019607843137251!black},
y grid style={white!69.019607843137251!black}
]
\addplot graphics [includegraphics cmd=\pgfimage,xmin=0, xmax=1, ymin=0, ymax=1] {Fig_porto_3d1.png};
\addplot graphics [includegraphics cmd=\pgfimage,xmin=0, xmax=1, ymin=0, ymax=1] {Fig_porto_3d2.png};
\node at (axis cs:0.99,0.01)[
  scale=0.5,
  anchor=south east,
  text=black,
  rotate=0.0
]{  24:00};
\end{groupplot}

\end{tikzpicture}